\pdfoutput=1

\documentclass[11pt]{article}

\usepackage[preprint]{acl}
\usepackage{amsmath}
\usepackage{amssymb}
\usepackage{times}
\usepackage{latexsym}
\usepackage{enumitem}
\usepackage{booktabs}
\usepackage{multirow}
\usepackage{bbm}

\usepackage[T1]{fontenc}

\usepackage[utf8]{inputenc}

\usepackage{microtype}

\usepackage{inconsolata}

\usepackage{graphicx}

\usepackage[most]{tcolorbox}
\usepackage{xcolor}
\usepackage{varwidth}  
\usepackage{hyperref}
\definecolor{Link}{HTML}{FFFFFF}
\tcbset{acad/base/.style={
  enhanced, breakable,
  boxsep=5pt, left=8pt, right=8pt, top=6pt, bottom=6pt, colback=white,
  fonttitle=\bfseries,
   before title={\begingroup\hypersetup{
      linkcolor=Link,
  }},
  after title={\endgroup}
}}

\title{Does Localization Inform Unlearning? A Rigorous Examination of Local Parameter Attribution for Knowledge Unlearning in Language Models}

\author{Hwiyeong Lee \: Uiji Hwang \: Hyelim Lim \: Taeuk Kim\thanks{Corresponding author} \\[0.5ex]
  Hanyang University, Seoul, Republic of Korea \\[0.5ex]
  \texttt{\{hyglee,willpower,yomilimi,kimtaeuk\}@hanyang.ac.kr}}

\begin{document}
\maketitle
\begin{abstract}

Large language models often retain unintended content, prompting growing interest in knowledge unlearning.
Recent approaches emphasize localized unlearning, restricting parameter updates to specific regions in an effort to remove target knowledge while preserving unrelated general knowledge. 
However, their effectiveness remains uncertain due to the lack of robust and thorough evaluation of the trade-off between the competing goals of unlearning.
In this paper, we begin by revisiting existing localized unlearning approaches. We then conduct controlled experiments to rigorously evaluate whether local parameter updates causally contribute to unlearning.
Our findings reveal that the set of parameters that must be modified for effective unlearning is not strictly determined, challenging the core assumption of localized unlearning that parameter locality is inherently indicative of effective knowledge removal. We release our code at \url{https://github.com/HYU-NLP/loc-unlearn}

\end{abstract}

\section{Introduction}

Due to large-scale pretraining, large language models (LLMs) often internalize not only useful knowledge but also harmful biases, sensitive data, and copyrighted or outdated content \cite{chang-etal-2023-speak, mozes2023usellmsillicitpurposes, eldan2023s, yeetaal}.
This has sparked growing interest in machine unlearning for LLMs, a post-training technique that selectively removes such information without full retraining \cite{blanco2025digital, liu2025rethinking}.
Despite their promise, current methods face key limitations: inadvertent forgetting of unrelated knowledge, susceptibility to prompt rephrasing, and vulnerability to information extraction under white-box conditions \cite{patil2023can, lynch2024eight}.


In response, recent research has incorporated the notion of localization \cite{hase2023does} into knowledge unlearning, aiming to first pinpoint parameter regions presumed to store the target knowledge, and subsequently confine unlearning updates to those regions \cite{tian2024forget, jia2024wagle, wang2024detoxifying}.
These studies commonly highlight that unrestrained parameter updates in unlearning lead to undesirable forgetting of general knowledge, and that unlearning should instead target a critical subset of weights, thereby preserving the model’s overall utility. 

While the idea of \textit{localized unlearning} is promising, we identify critical gaps that have largely been overlooked in this line of work. 
First, current approaches frequently depend on surface-level output evaluation metrics (e.g., ROUGE-L \cite{lin-2004-rouge}) to quantify the degree of knowledge embedded in model parameters; however, these metrics are recently acknowledged to be unreliable for such assessments \cite{hong2024intrinsic, wang2025towards}.

Second, fair comparison across existing methods is hindered by the inherent nature of unlearning, which requires balancing the removal of targeted knowledge with the retention of general utility.
This trade-off obscures clear comparisons, as different approaches often excel in different aspects of the unlearning task \cite{wang2025towards}.

Most notably, prior work on localized unlearning tends to emphasize the design of localization techniques while relying on the unverified assumption that parameter locality inherently reflects unlearning effectiveness, without establishing whether the identified regions play a causal role.
As a result, the underlying connection between localization and knowledge unlearning remains unexplored.

In this paper, we investigate whether the success of localization truly translates into improved unlearning, particularly by leveraging a controlled environment where the ground-truth parameter regions responsible for storing the target knowledge are explicitly predefined. 
This setup allows us to disentangle the contribution of localization to unlearning, rather than evaluating localization itself. 

Our findings are surprising: even when unlearning is performed on the ground-truth region, it does not necessarily yield a better trade-off between forgetting and retention. 
This challenges the core assumption underlying localized unlearning that constraining parameter updates to specific regions helps preserve unrelated knowledge elsewhere in the model. 
Ultimately, we question the traditional view of unlearning as full parameter recovery, suggesting that the set of parameters to be updated is not strictly given, and that the model may achieve ideal unlearning via flexible parameter adaptation.

\section{Background and Related Work}
\paragraph{LLM unlearning methods}
Most unlearning approaches rely on fine-tuning the target model, with objectives falling into three categories: (1) gradient ascent, which minimizes the model's likelihood on sentences encoding the target knowledge; (2) preference optimization, which treats target knowledge as negative examples; and (3) representation learning, which randomizes internal representations for inputs containing the target knowledge. To represent these paradigms, we evaluate four methods: WGA \cite{wang2025rethinking}, NPO \cite{zhang2024negative}, DPO \cite{rafailov2023direct}, and RMU \cite{li2024the}, with details in Appendix \ref{sec:appendixA}.


\paragraph{Knowledge storage in LLMs}
\label{sec:keyvalue}
A growing body of work in the field of mechanistic interpretability suggests that in Transformer-based LLMs, multilayer perceptrons (MLPs) play a crucial role in storing factual knowledge \cite{meng2022locating, geva-etal-2021-transformer, geva-etal-2022-transformer}. Specifically, \citet{geva-etal-2021-transformer} propose that MLPs can be understood as emulated key-value memories \cite{sukhbaatar2015end}: the first linear layer projects input hidden states into a latent key space to produce memory coefficients, while the second layer maps these coefficients to value vectors that encode factual information.

Formally, an MLP in the $\ell$-th Transformer layer accepts a hidden state $\mathbf{x}^\ell \in \mathbb{R}^{d_m}$ as input and processes it through two linear layers with a non-linearity $f(\cdot)$ in between. The final MLP output is computed as:
 \begin{equation*}
  \label{eq:mlp_combined}
  \mathbf{M}^\ell = f(W_K^\ell \mathbf{x}^\ell) W_V^\ell = \sum_{i=1}^{d_{\text{ff}}} m_i^\ell \cdot \mathbf{v}_i^\ell,
\end{equation*}
where $W_K^\ell \in \mathbb{R}^{d_{\text{ff}} \times d_m}$ and $W_V^\ell \in \mathbb{R}^{d_{\text{ff}} \times d_m}$ are the weight matrices of the MLP's first and second linear layers, respectively. The intermediate activations $\mathbf{m}^\ell = f(W_K^\ell \mathbf{x}^\ell)$ serve as memory coefficients, and $\mathbf{v}_i^\ell \in \mathbb{R}^{d_m}$, the $i$-th row of $W_V^\ell$, is referred to as a \textit{value vector}. This formulation allows the MLP output to be interpreted as a linear combination of value vectors, each weighted by its corresponding memory coefficient.

In the context of knowledge unlearning, \citet{hong-etal-2024-dissecting} highlight the need for unlearning techniques that effectively modify the value vectors where knowledge is stored, showing that current methods induce modifications in the knowledge retrieval process rather than the value vectors themselves. 
Building on this insight, our investigation into localization for knowledge unlearning attributes factual knowledge to a specific set of value vectors and designates them as the target components for localization, allowing us to probe whether localization offers a viable path forward for addressing this challenge.

\paragraph{Localization}
Localization is broadly defined as the task of identifying the components of a model responsible for specific knowledge or behavior \cite{hase2023does}. 
This notion has been widely adopted in the field of model editing, particularly within the locate-then-edit paradigm \cite{meng2022locating, meng2023massediting}. 
Aligned with this line of work, recent studies in knowledge unlearning increasingly incorporate localization techniques \cite{jia2024wagle, tian2024forget}. Yet, their performance gains remain questionable given the lack of robustness in unlearning evaluations, and the supposed informativeness of localization for unlearning remains a tenuous assumption requiring rigorous validation.

Meanwhile, this is not the first work to scrutinize the causal validity of localization. 
Notably, \citet{hase2023does} examine whether Causal Tracing \cite{meng2022locating} aids factual knowledge editing, and \citet{wang2024does} evaluate Inference-Time Intervention \cite{li2023inferencetime} in steering a model’s truthful behavior. 
However, these studies are limited to testing a specific localization method in the context of single-shot editing. 
In contrast, our key contribution is to adopt a method-agnostic perspective that isolates and evaluates the causal impact of localization success on fine-tuning–based knowledge unlearning.


\section{Revisiting Localized Unlearning} \label{sec:preliminaries}
\paragraph{Datasets and models}
In this paper, we conduct experiments using the TOFU dataset \cite{maini2024tofu}, which is widely adopted in the field of unlearning research. It consists of 4,000 synthetic QA pairs about fictitious authors, for which we employ a split of 10\% as the forget set and the remaining 90\% as the retain set in our experimental setup.
We consider two recent open-source LLMs: LLaMA3.1-8B-Instruct \cite{grattafiori2024llama} and OLMo2-7B-Instruct \cite{olmo20242}.

\paragraph{Unlearning evaluation}
Knowledge unlearning aims to achieve two primary objectives: the removal of target knowledge and the preservation of the rest \cite{jang-etal-2023-knowledge, si2023knowledge}. 
To enable a comprehensive and robust evaluation of unlearning methods, we decompose this goal into two components: (1) quantifying the extent of knowledge parameterization, and (2) enabling a fair comparison of trade-offs between forgetting and retention.

Regarding (1), we not only adopt Forget Quality (FQ) and Model Utility (MU) as provided by TOFU, but also incorporate \textit{Extraction Strength} (ES) \cite{DBLP:journals/corr/abs-2012-07805}. 
ES has recently been proposed as a robust alternative \cite{wang2025towards} to traditional metrics such as Perplexity \cite{chang2024survey} and ROUGE-L, which have been criticized for their limited capacity to capture the internalized knowledge embedded in model parameters \cite{hong2024intrinsic, wang2025towards}. 
ES is computed on forget and retain sets, defining \textit{Forget Strength (FS)} as $1 - \text{ES}_{\text{forget}}$ and \textit{Retain Strength (RS)} as $\text{ES}_{\text{retain}}$, where higher values indicate stronger forgetting and retention, respectively. 

Regarding (2), we note that prior works often report unlearning performance at a single point along the unlearning process. 
However, unlearning typically entails a trade-off: as the model increasingly forgets the target knowledge, its ability to retain general utility tends to decline. 
Therefore, comparing different methods at an arbitrary point in this process can be misleading, as each method may favor a different side of the trade-off. 

To this end, we adopt two evaluation strategies, following practices from Out-of-Distribution Detection research. 
First, we perform a controlled single-point comparison, denoted as \textit{MU95}, by measuring FQ at the point where MU reaches 95\% of the target model’s initial value. This design ensures a fair comparison across methods by standardizing the retention level to a tolerable degree of degradation. 
Second, to evaluate whether unlearning methods consistently guide the model toward more desirable parameter updates throughout the process, we compute the area under the FS–RS curve, referred to as \textit{AUES} (Area Under the Extraction Strength curve). AUES captures the overall trade-off between forgetting and retention over the unlearning trajectory.
Details of the evaluation are provided in Appendix \ref{sec:appendixB}.


\begin{table}[t]
  \centering
  \scriptsize
  \renewcommand{\arraystretch}{1.2}
  \begin{tabular}{llc@{\hspace{1.0em}}c}
    \toprule
    \textbf{Model} & \textbf{Method} & \textbf{AUES} (\textuparrow) & \textbf{MU95} (\textuparrow) \\
    \midrule
    \multirow{5}{*}{LLaMA3.1-8B-Instruct} 
      & Original & -- & -20.37 \\
      \cmidrule{2-4}
      & Random      & $\textbf{0.529}_{\pm \text{0.005}}$ & $\textbf{-14.87}_{\pm \text{0.33}}$ \\
      & Activations & 0.522 & -16.84 \\
      & MemFlex     & 0.491 & -15.97 \\
      & WAGLE       & 0.525 & -16.61 \\
    \midrule
    \multirow{5}{*}{OLMo2-7B-Instruct} 
      & Original & -- & -21.10 \\
      \cmidrule{2-4}
      & Random      & $\textbf{0.582}_{\pm \text{0.001}}$ & $\textbf{-14.13}_{\pm \text{0.22}}$ \\
      & Activations & 0.542 & -14.44 \\
      & MemFlex     & 0.508 & -16.53 \\
      & WAGLE       & 0.517 & -15.17 \\
    \bottomrule
  \end{tabular}
  \caption{Comparison of AUES and MU95 for different localization methods. 
  Higher AUES and MU95 indicate better trade-off between forgetting and retention. `Original' denotes the state of the model before unlearning. For the `Random' baseline, results are averaged over three random seeds. The best scores are in \textbf{bold}.}
  \label{tab:preliminary}
\end{table}
\paragraph{Revisiting current approaches}
\label{revisit}

Using the evaluation framework above, we measure existing localized unlearning approaches, including Activations \cite{chang-etal-2024-localization}, MemFlex \cite{tian2024forget}, and WAGLE \cite{jia2024wagle}. 
For each method, we follow its proposed localization strategies to score value vectors by relevance to the target knowledge. 
We apply the NPO objective to the top 10\% of ranked vectors and compare the results against randomly selected vectors of the same size. Details of the methods are presented in Appendix \ref{sec:appendixC}.

As illustrated in Table~\ref{tab:preliminary}, we observe that unlearning over randomly selected regions outperforms unlearning over the regions selected by localization methods. 
While this result points to the failure of current localized unlearning approaches, it raises a deeper question: \textit{is this simply a limitation of existing localization strategies, or does it cast doubt on the very existence of a solution that the localization seeks to uncover?} 
This motivates the investigations in the following section.
\section{Controlled Experiments}
\label{controlled_exp}
\begin{table*}[t]
    \centering
    \small
    \renewcommand{\arraystretch}{1.2}
    \setlength{\tabcolsep}{0.5em}
        \begin{tabular}{cccccccccc}
        \toprule
        \multirow{2}{*}{\textbf{Method}} & \multirow{2}{*}{\textbf{Metric}} 
          & \multicolumn{4}{c}{\textbf{LLaMA3.1-8B-Instruct}} & \multicolumn{4}{c}{\textbf{OLMo2-7B-Instruct}} \\
        \cmidrule(lr){3-6} \cmidrule(lr){7-10}
        & & Random & Oracle & $|\Delta|$ & $p$-value 
          & Random & Oracle & $|\Delta|$ & $p$-value \\
        \midrule
        \multirow{2}{*}{WGA}
          & AUES (\textuparrow)  & $0.586_{\pm \text{0.020}}$ & $\textbf{0.593}_{\pm \text{0.016}}$ & $0.018_{\pm \text{0.013}}$ & 0.61
                             & $\textbf{0.609}_{\pm \text{0.020}}$ & $0.605_{\pm \text{0.011}}$ & $0.008_{\pm \text{0.007}}$ & 0.64 \\
          & MU95 (\textuparrow) & $\text{-}10.33_{\pm \text{1.73}}$ & $\textbf{\text{-}10.00}_{\pm \text{0.42}}$ & $0.86_{\pm \text{1.02}}$   & 0.46 
                             & $\text{-}13.88_{\pm \text{0.58}}$ & $\textbf{\text{-}13.77}_{\pm \text{0.45}}$ & $0.23_{\pm \text{0.20}}$   & 0.56 \\
        \midrule
        \multirow{2}{*}{NPO}
          & AUES (\textuparrow)  & $\textbf{0.625}_{\pm \text{0.016}}$ & $0.619_{\pm \text{0.011}}$ & $0.011_{\pm \text{0.011}}$ & 0.71
                             & $0.638_{\pm \text{0.015}}$ & $\textbf{0.639}_{\pm \text{0.017}}$ & $0.007_{\pm \text{0.004}}$ & 0.52  \\
          & MU95 (\textuparrow) & $\text{-}9.45_{\pm \text{1.91}}$  & $\textbf{\text{-}8.56}_{\pm \text{1.39}}$  & $0.90_{\pm \text{0.66}}$   & 0.31
                             & $\textbf{\text{-}14.19}_{\pm \text{0.47}}$ & $\text{-}14.33_{\pm \text{0.58}}$ & $0.14_{\pm \text{0.12}}$   & 0.72 \\
        \midrule
        \multirow{2}{*}{DPO}
          & AUES (\textuparrow)  & $\textbf{0.497}_{\pm \text{0.013}}$ & $0.492_{\pm \text{0.012}}$ & $0.007_{\pm \text{0.008}}$ & 0.66 
                             & $0.561_{\pm \text{0.011}}$ & $\textbf{0.568}_{\pm \text{0.013}}$ & $0.010_{\pm \text{0.008}}$ & 0.36 \\
          & MU95 (\textuparrow)& $\textbf{\text{-}13.26}_{\pm \text{1.06}}$ & $\text{-}13.60_{\pm \text{0.72}}$ & $1.09_{\pm \text{1.06}}$   & 0.68
                             & $\text{-}13.62_{\pm \text{0.31}}$ & $\textbf{\text{-}13.52}_{\pm \text{0.53}}$ & $0.41_{\pm \text{0.17}}$   & 0.57 \\
        \midrule
        \multirow{2}{*}{RMU}
          & AUES (\textuparrow)  & $\textbf{0.506}_{\pm \text{0.030}}$ & $0.502_{\pm \text{0.017}}$ & $0.017_{\pm \text{0.010}}$ & 0.37 
                             & $0.437_{\pm \text{0.024}}$ & $\textbf{0.439}_{\pm \text{0.020}}$ & $0.004_{\pm \text{0.003}}$ & 0.62 \\
          & MU95 (\textuparrow)& $\text{-}13.75_{\pm \text{0.91}}$ & $\textbf{\text{-}13.64}_{\pm \text{0.63}}$ & $0.45_{\pm \text{0.24}}$   & 0.39
                             & $\textbf{\text{-}12.95}_{\pm \text{0.69}}$ & $\text{-}13.62_{\pm \text{1.32}}$ & $0.68_{\pm \text{0.79}}$   & 0.57 \\
        \bottomrule
    \end{tabular}
    \caption{
    Comparison of AUES and MU95 between Random and Oracle scenarios across two different LLMs. 
    $|\Delta|$ denotes the absolute difference in scores between the two settings, while $p$-value indicates the statistical significance of this difference.
    For each method, we report the mean with the standard deviation as a subscript, computed across five random seeds. 
    Details of the $p$-value computation are provided in Appendix \ref{sec:appendixD}. 
    The best scores are in \textbf{bold}.
    }
    \label{tab:main}
\end{table*}
\paragraph{Experimental design}
In this part, we aim to examine whether localization truly provides a distinctive basis for guiding knowledge unlearning. 
While the previous experiment in \S \ref{sec:preliminaries} underscores the potential ineffectiveness of localization, this outcome could be attributed not to the causal link between localization and unlearning, but rather to a failure of the localization process itself. 
In other words, given the incompleteness of current localization methods \cite{chang-etal-2024-localization}, the result may simply reflect that these approaches fail to identify the appropriate parameter region. 
To decouple and eliminate localization accuracy as a confounding factor, we design a controlled experiment where the ground-truth region is explicitly predefined, allowing us to assume perfect localization. 
The specific operation process is as follows:
\begin{enumerate}[leftmargin=12pt, itemsep=1pt, topsep=5pt]
    \item We begin by fine-tuning a pretrained model \( \theta_p \) on the retain set only, using all model parameters, and obtain the resulting model \( \theta_r \). Note that \( \theta_r \) serves as the gold standard in unlearning.
    \item We randomly select \( p\% \) of the value vectors from the entire model and define them as the \textbf{target region}, denoted \( \mathcal{V}_{\text{tgt}} \). We then train \( \theta_r \) on the forget set, applying updates only to the value vectors in \(\mathcal{V}_{\text{tgt}}\), yielding \( \theta_o \). This ensures that learning effects on the forget set remain confined to the target region, allowing us to fully attribute target knowledge to the value vectors in \( \mathcal{V}_{\text{tgt}} \).
    \item Again, we randomly select another \( p\% \) of value vectors from outside the target region, i.e., from \( \mathcal{V} \setminus \mathcal{V}_{\text{tgt}} \), and define this as the \textbf{random region}, denoted as \( \mathcal{V}_{\text{rdm}} \). We then perform unlearning from \( \theta_o \) using updates restricted to the value vectors in \( \mathcal{V}_{\text{tgt}} \) and \( \mathcal{V}_{\text{rdm}} \), denoting each scenario as \textbf{Oracle} and \textbf{Random}, respectively.
\end{enumerate}
As the goal of localization is to identify \( \mathcal{V}_{\text{tgt}} \), \textbf{Oracle} simulates an idealized localization, while \textbf{Random} serves as its comparative counterpart. 
By comparing the two, we examine whether localization acts as a necessary condition for effective unlearning. 
The proportion $p$ is set to 10\%, as it offers a good compromise—large enough to allow learning from the forget data, yet small enough to maintain a meaningful degree of locality.

\paragraph{Results}
From Table \ref{tab:main}, we observe a surprising result: the improvement offered by Oracle over Random is marginal (with all $p$-values exceeding 0.3) and, in some cases, Random even outperforms Oracle. 
This trend consistently holds across different model types and unlearning methods.

We regard this as compelling evidence against the assumption that a fixed set of parameters must be updated to achieve effective unlearning. 
The results suggest that unlearning may not rely on a specific parameter region, but can instead be achieved through multiple alternative regions in the model.

\paragraph{Further investigation of unlearning objectives}

While our findings are significant, we take a further step to examine whether this observation is merely a consequence of the limitations of current unlearning approaches, all of which operate at the output level. 
That is, rather than explicitly specifying the target values that the value vectors in \( \mathcal{V}_{\text{tgt}} \) should aim to reach, existing methods typically rely on fine-tuning, adjusting model parameters by minimizing a loss computed over the final outputs. 
Notably, localized unlearning is grounded in the assumption that the goal of unlearning is to revert the model back to  \( \theta_r \), thereby placing emphasis on identifying which parameters should be updated—ideally, those in \( \mathcal{V}_{\text{tgt}} \). 
However, when supervision is indirect, optimization may permit diverse parameter configurations within \( \mathcal{V}_{\text{tgt}} \) that still satisfy the objective. 
As a result, the model may fail to fully leverage the benefits of localization, leading to underutilization of informative signals.

\begin{figure}[t]
  \includegraphics[width=\columnwidth]{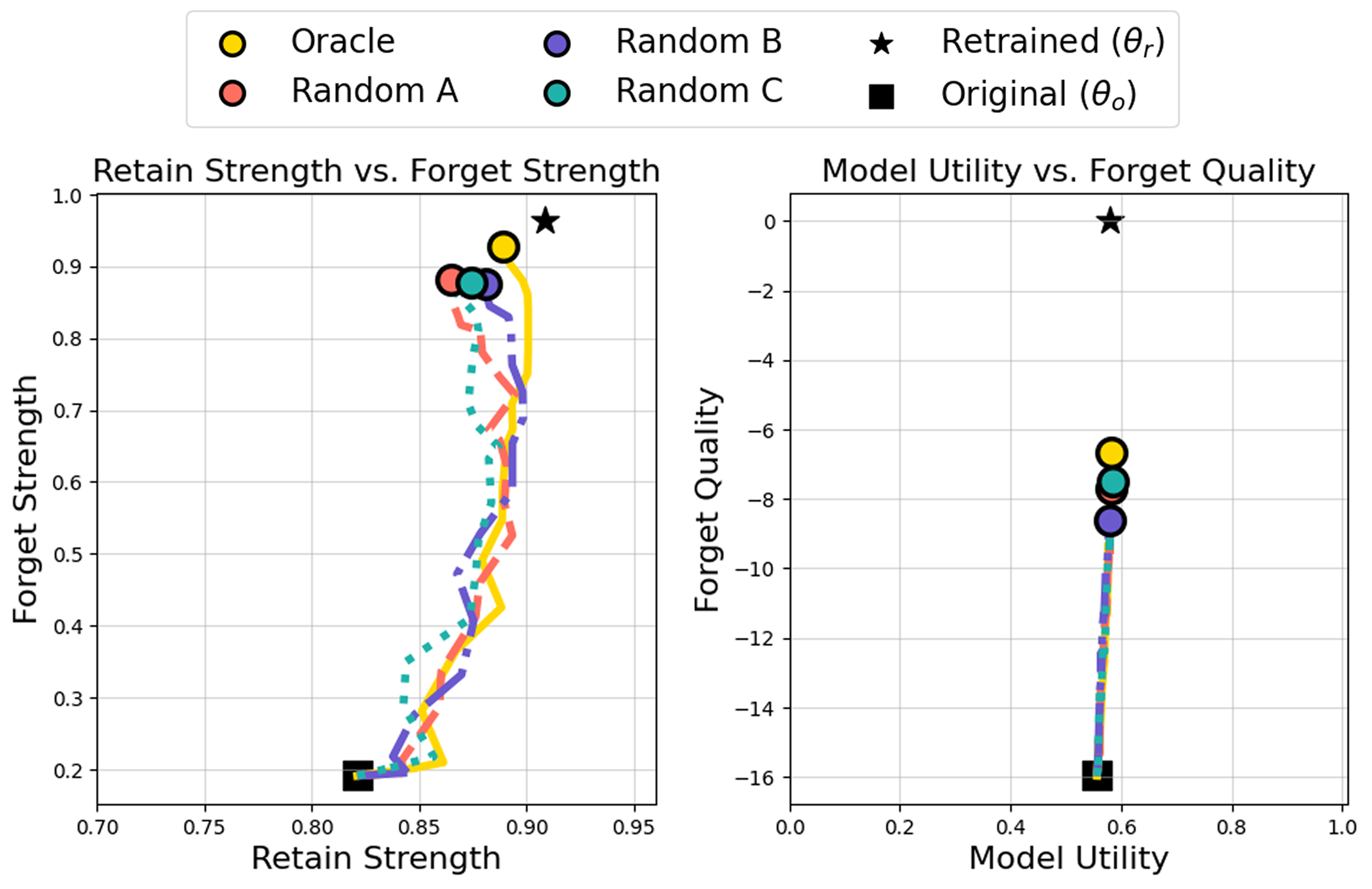}
  \caption{ Comparison of Oracle and Random scenarios updated via L2 minimization of the MLP outputs at each layer. Plots show RS vs. FS (left) and MU vs. FQ (right), conducted on the LLaMA3.1-8B-Instruct.}
  \label{fig:discussion}
\end{figure}

To this end, we revisit the Oracle vs. Random experiment using an alternative unlearning mechanism that more \textit{directly} supervises parameter updates: instead of relying on output-level signals, we minimize the L2 distance between the MLP outputs at each layer and those produced by \( \theta_r \). 
The results shown in Figure~\ref{fig:discussion} are remarkable: even when adjusting \( \mathcal{V}_{\text{rdm}} \), the MLP outputs of \( \theta_r \) can be reproduced to a degree comparable to \( \mathcal{V}_{\text{tgt}} \). 
This raises a critical question of whether the set of value vectors to be edited is strictly confined to \( \mathcal{V}_{\text{tgt}} \), or if resembling \( \theta_r \) can be achieved through flexible adaptation of alternative regions, such as \( \mathcal{V}_{\text{rdm}} \).



\section{Conclusion}
In this paper, we have rigorously examined whether localization truly provides an effective basis for unlearning. 
We begin by proposing an improved framework to address shortcomings in unlearning evaluation. 
Prompted by the breakdown of existing methods under this framework, we conduct controlled experiments suggesting that the failure of localized unlearning may stem from the absence of a uniquely responsible parameter region.


\section*{Limitations}
We follow prior work \cite{geva-etal-2021-transformer, geva-etal-2022-transformer, geva-etal-2023-dissecting, meng2022locating, chang-etal-2024-localization, hong-etal-2024-dissecting}  in assuming that MLPs are the primary components in LLMs responsible for storing knowledge. Accordingly, we restrict our localization analysis to MLPs and do not consider other components such as attention layers. To handle value vectors within MLPs as the unit of localization and enable fair comparisons across methods, we reformulate MemFlex \cite{tian2024forget} and WAGLE \cite{jia2024wagle} to score value vectors, as each originally defines the localization unit differently—individual weights in WAGLE and LoRA modules in MemFlex (see Appendix \ref{sec:appendixC}). This modification may not precisely reflect the original design intentions of each method. To assess the impact of this choice, we additionally report complementary results in Appendix \ref{sec:appendixE}.

In \S \ref{controlled_exp}, to restrict the influence of the forget data to a predetermined set of value vectors (i.e., the target region), we trained the model by updating only that region while freezing the rest. However, this is a controlled experimental setup rather than a realistic scenario, and it remains unclear whether the findings generalize to models trained with updates applied to all parameters. We roughly address this issue by searching for an optimal injection ratio (i.e., 10\%) that maintains comparable training performance to full-parameter updates, under the assumption that such a model would generalize better to the full-parameter setting.


\section*{Acknowledgments}

This work was supported by the Institute of Information \& communications Technology Planning \& evaluation (IITP) grant funded by the Korea government (MSIT) (RS-2020-II201373, Artificial Intelligence Graduate School Program (Hanyang University); IITP-2025-RS-2023-00253914, Artificial Intelligence Semiconductor Support Program to nurture the best talents), and by the National Research Foundation of Korea (NRF) grant funded by the Korea government (MSIT) (RS-2025-00558151).

\bibliography{custom}

\begin{thebibliography}{39}
\providecommand{\natexlab}[1]{#1}

\bibitem[{Blanco-Justicia et~al.(2025)Blanco-Justicia, Jebreel, Manzanares-Salor, S{\'a}nchez, Domingo-Ferrer, Collell, and Eeik~Tan}]{blanco2025digital}
Alberto Blanco-Justicia, Najeeb Jebreel, Benet Manzanares-Salor, David S{\'a}nchez, Josep Domingo-Ferrer, Guillem Collell, and Kuan Eeik~Tan. 2025.
\newblock Digital forgetting in large language models: A survey of unlearning methods.
\newblock \emph{Artificial Intelligence Review}, 58(3):90.

\bibitem[{Carlini et~al.(2020)Carlini, Tram{\`{e}}r, Wallace, Jagielski, Herbert{-}Voss, Lee, Roberts, Brown, Song, Erlingsson, Oprea, and Raffel}]{DBLP:journals/corr/abs-2012-07805}
Nicholas Carlini, Florian Tram{\`{e}}r, Eric Wallace, Matthew Jagielski, Ariel Herbert{-}Voss, Katherine Lee, Adam Roberts, Tom~B. Brown, Dawn Song, {\'{U}}lfar Erlingsson, Alina Oprea, and Colin Raffel. 2020.
\newblock \href {https://arxiv.org/abs/2012.07805} {Extracting training data from large language models}.
\newblock \emph{CoRR}, abs/2012.07805.

\bibitem[{Chang et~al.(2023)Chang, Cramer, Soni, and Bamman}]{chang-etal-2023-speak}
Kent Chang, Mackenzie Cramer, Sandeep Soni, and David Bamman. 2023.
\newblock \href {https://doi.org/10.18653/v1/2023.emnlp-main.453} {Speak, memory: An archaeology of books known to {C}hat{GPT}/{GPT}-4}.
\newblock In \emph{Proceedings of the 2023 Conference on Empirical Methods in Natural Language Processing}, pages 7312--7327, Singapore. Association for Computational Linguistics.

\bibitem[{Chang et~al.(2024{\natexlab{a}})Chang, Thomason, and Jia}]{chang-etal-2024-localization}
Ting-Yun Chang, Jesse Thomason, and Robin Jia. 2024{\natexlab{a}}.
\newblock \href {https://doi.org/10.18653/v1/2024.naacl-long.176} {Do localization methods actually localize memorized data in {LLM}s? a tale of two benchmarks}.
\newblock In \emph{Proceedings of the 2024 Conference of the North American Chapter of the Association for Computational Linguistics: Human Language Technologies (Volume 1: Long Papers)}, pages 3190--3211, Mexico City, Mexico. Association for Computational Linguistics.

\bibitem[{Chang et~al.(2024{\natexlab{b}})Chang, Wang, Wang, Wu, Yang, Zhu, Chen, Yi, Wang, Wang et~al.}]{chang2024survey}
Yupeng Chang, Xu~Wang, Jindong Wang, Yuan Wu, Linyi Yang, Kaijie Zhu, Hao Chen, Xiaoyuan Yi, Cunxiang Wang, Yidong Wang, and 1 others. 2024{\natexlab{b}}.
\newblock A survey on evaluation of large language models.
\newblock \emph{ACM transactions on intelligent systems and technology}, 15(3):1--45.

\bibitem[{Eldan and Russinovich(2023)}]{eldan2023s}
Ronen Eldan and Mark Russinovich. 2023.
\newblock Who's harry potter? approximate unlearning in llms.
\newblock \emph{arXiv preprint arXiv:2310.02238}.

\bibitem[{Geva et~al.(2023)Geva, Bastings, Filippova, and Globerson}]{geva-etal-2023-dissecting}
Mor Geva, Jasmijn Bastings, Katja Filippova, and Amir Globerson. 2023.
\newblock \href {https://doi.org/10.18653/v1/2023.emnlp-main.751} {Dissecting recall of factual associations in auto-regressive language models}.
\newblock In \emph{Proceedings of the 2023 Conference on Empirical Methods in Natural Language Processing}, pages 12216--12235, Singapore. Association for Computational Linguistics.

\bibitem[{Geva et~al.(2022)Geva, Caciularu, Wang, and Goldberg}]{geva-etal-2022-transformer}
Mor Geva, Avi Caciularu, Kevin Wang, and Yoav Goldberg. 2022.
\newblock \href {https://doi.org/10.18653/v1/2022.emnlp-main.3} {Transformer feed-forward layers build predictions by promoting concepts in the vocabulary space}.
\newblock In \emph{Proceedings of the 2022 Conference on Empirical Methods in Natural Language Processing}, pages 30--45, Abu Dhabi, United Arab Emirates. Association for Computational Linguistics.

\bibitem[{Geva et~al.(2021)Geva, Schuster, Berant, and Levy}]{geva-etal-2021-transformer}
Mor Geva, Roei Schuster, Jonathan Berant, and Omer Levy. 2021.
\newblock \href {https://doi.org/10.18653/v1/2021.emnlp-main.446} {Transformer feed-forward layers are key-value memories}.
\newblock In \emph{Proceedings of the 2021 Conference on Empirical Methods in Natural Language Processing}, pages 5484--5495, Online and Punta Cana, Dominican Republic. Association for Computational Linguistics.

\bibitem[{Grattafiori et~al.(2024)Grattafiori, Dubey, Jauhri, Pandey, Kadian, Al-Dahle, Letman, Mathur, Schelten, Vaughan et~al.}]{grattafiori2024llama}
Aaron Grattafiori, Abhimanyu Dubey, Abhinav Jauhri, Abhinav Pandey, Abhishek Kadian, Ahmad Al-Dahle, Aiesha Letman, Akhil Mathur, Alan Schelten, Alex Vaughan, and 1 others. 2024.
\newblock The llama 3 herd of models.
\newblock \emph{arXiv preprint arXiv:2407.21783}.

\bibitem[{Hase et~al.(2023)Hase, Bansal, Kim, and Ghandeharioun}]{hase2023does}
Peter Hase, Mohit Bansal, Been Kim, and Asma Ghandeharioun. 2023.
\newblock Does localization inform editing? surprising differences in causality-based localization vs. knowledge editing in language models.
\newblock \emph{Advances in Neural Information Processing Systems}, 36:17643--17668.

\bibitem[{Hong et~al.(2024{\natexlab{a}})Hong, Yu, Yang, Ravfogel, and Geva}]{hong2024intrinsic}
Yihuai Hong, Lei Yu, Haiqin Yang, Shauli Ravfogel, and Mor Geva. 2024{\natexlab{a}}.
\newblock Intrinsic evaluation of unlearning using parametric knowledge traces.
\newblock \emph{arXiv preprint arXiv:2406.11614}.

\bibitem[{Hong et~al.(2024{\natexlab{b}})Hong, Zou, Hu, Zeng, Wang, and Yang}]{hong-etal-2024-dissecting}
Yihuai Hong, Yuelin Zou, Lijie Hu, Ziqian Zeng, Di~Wang, and Haiqin Yang. 2024{\natexlab{b}}.
\newblock \href {https://doi.org/10.18653/v1/2024.emnlp-main.228} {Dissecting fine-tuning unlearning in large language models}.
\newblock In \emph{Proceedings of the 2024 Conference on Empirical Methods in Natural Language Processing}, pages 3933--3941, Miami, Florida, USA. Association for Computational Linguistics.

\bibitem[{Jang et~al.(2023)Jang, Yoon, Yang, Cha, Lee, Logeswaran, and Seo}]{jang-etal-2023-knowledge}
Joel Jang, Dongkeun Yoon, Sohee Yang, Sungmin Cha, Moontae Lee, Lajanugen Logeswaran, and Minjoon Seo. 2023.
\newblock \href {https://doi.org/10.18653/v1/2023.acl-long.805} {Knowledge unlearning for mitigating privacy risks in language models}.
\newblock In \emph{Proceedings of the 61st Annual Meeting of the Association for Computational Linguistics (Volume 1: Long Papers)}, pages 14389--14408, Toronto, Canada. Association for Computational Linguistics.

\bibitem[{Jia et~al.(2024)Jia, Liu, Zhang, Ram, Baracaldo, and Liu}]{jia2024wagle}
Jinghan Jia, Jiancheng Liu, Yihua Zhang, Parikshit Ram, Nathalie Baracaldo, and Sijia Liu. 2024.
\newblock Wagle: Strategic weight attribution for effective and modular unlearning in large language models.
\newblock \emph{arXiv preprint arXiv:2410.17509}.

\bibitem[{Lhoest et~al.(2021)Lhoest, Villanova~del Moral, Jernite, Thakur, von Platen, Patil, Chaumond, Drame, Plu, Tunstall, Davison, {\v{S}}a{\v{s}}ko, Chhablani, Malik, Brandeis, Le~Scao, Sanh, Xu, Patry, McMillan-Major, Schmid, Gugger, Delangue, Matussi{\`e}re, Debut, Bekman, Cistac, Goehringer, Mustar, Lagunas, Rush, and Wolf}]{lhoest-etal-2021-datasets}
Quentin Lhoest, Albert Villanova~del Moral, Yacine Jernite, Abhishek Thakur, Patrick von Platen, Suraj Patil, Julien Chaumond, Mariama Drame, Julien Plu, Lewis Tunstall, Joe Davison, Mario {\v{S}}a{\v{s}}ko, Gunjan Chhablani, Bhavitvya Malik, Simon Brandeis, Teven Le~Scao, Victor Sanh, Canwen Xu, Nicolas Patry, and 13 others. 2021.
\newblock \href {https://doi.org/10.18653/v1/2021.emnlp-demo.21} {Datasets: A community library for natural language processing}.
\newblock In \emph{Proceedings of the 2021 Conference on Empirical Methods in Natural Language Processing: System Demonstrations}, pages 175--184, Online and Punta Cana, Dominican Republic. Association for Computational Linguistics.

\bibitem[{Li et~al.(2023)Li, Patel, Vi{\'e}gas, Pfister, and Wattenberg}]{li2023inferencetime}
Kenneth Li, Oam Patel, Fernanda Vi{\'e}gas, Hanspeter Pfister, and Martin Wattenberg. 2023.
\newblock \href {https://openreview.net/forum?id=aLLuYpn83y} {Inference-time intervention: Eliciting truthful answers from a language model}.
\newblock In \emph{Thirty-seventh Conference on Neural Information Processing Systems}.

\bibitem[{Li et~al.(2024)Li, Pan, Gopal, Yue, Berrios, Gatti, Li, Dombrowski, Goel, Mukobi, Helm-Burger, Lababidi, Justen, Liu, Chen, Barrass, Zhang, Zhu, Tamirisa, Bharathi, Herbert-Voss, Breuer, Zou, Mazeika, Wang, Oswal, Lin, Hunt, Tienken-Harder, Shih, Talley, Guan, Steneker, Campbell, Jokubaitis, Basart, Fitz, Kumaraguru, Karmakar, Tupakula, Varadharajan, Shoshitaishvili, Ba, Esvelt, Wang, and Hendrycks}]{li2024the}
Nathaniel Li, Alexander Pan, Anjali Gopal, Summer Yue, Daniel Berrios, Alice Gatti, Justin~D. Li, Ann-Kathrin Dombrowski, Shashwat Goel, Gabriel Mukobi, Nathan Helm-Burger, Rassin Lababidi, Lennart Justen, Andrew~Bo Liu, Michael Chen, Isabelle Barrass, Oliver Zhang, Xiaoyuan Zhu, Rishub Tamirisa, and 27 others. 2024.
\newblock \href {https://openreview.net/forum?id=xlr6AUDuJz} {The {WMDP} benchmark: Measuring and reducing malicious use with unlearning}.
\newblock In \emph{Forty-first International Conference on Machine Learning}.

\bibitem[{Lin(2004)}]{lin-2004-rouge}
Chin-Yew Lin. 2004.
\newblock \href {https://aclanthology.org/W04-1013/} {{ROUGE}: A package for automatic evaluation of summaries}.
\newblock In \emph{Text Summarization Branches Out}, pages 74--81, Barcelona, Spain. Association for Computational Linguistics.

\bibitem[{Liu et~al.(2025)Liu, Yao, Jia, Casper, Baracaldo, Hase, Yao, Liu, Xu, Li et~al.}]{liu2025rethinking}
Sijia Liu, Yuanshun Yao, Jinghan Jia, Stephen Casper, Nathalie Baracaldo, Peter Hase, Yuguang Yao, Chris~Yuhao Liu, Xiaojun Xu, Hang Li, and 1 others. 2025.
\newblock Rethinking machine unlearning for large language models.
\newblock \emph{Nature Machine Intelligence}, pages 1--14.

\bibitem[{Lynch et~al.(2024)Lynch, Guo, Ewart, Casper, and Hadfield-Menell}]{lynch2024eight}
Aengus Lynch, Phillip Guo, Aidan Ewart, Stephen Casper, and Dylan Hadfield-Menell. 2024.
\newblock Eight methods to evaluate robust unlearning in llms.
\newblock \emph{arXiv preprint arXiv:2402.16835}.

\bibitem[{Maini et~al.(2024)Maini, Feng, Schwarzschild, Lipton, and Kolter}]{maini2024tofu}
Pratyush Maini, Zhili Feng, Avi Schwarzschild, Zachary~Chase Lipton, and J~Zico Kolter. 2024.
\newblock \href {https://openreview.net/forum?id=B41hNBoWLo} {{TOFU}: A task of fictitious unlearning for {LLM}s}.
\newblock In \emph{First Conference on Language Modeling}.

\bibitem[{Meng et~al.(2022)Meng, Bau, Andonian, and Belinkov}]{meng2022locating}
Kevin Meng, David Bau, Alex~J Andonian, and Yonatan Belinkov. 2022.
\newblock \href {https://openreview.net/forum?id=-h6WAS6eE4} {Locating and editing factual associations in {GPT}}.
\newblock In \emph{Advances in Neural Information Processing Systems}.

\bibitem[{Meng et~al.(2023)Meng, Sharma, Andonian, Belinkov, and Bau}]{meng2023massediting}
Kevin Meng, Arnab~Sen Sharma, Alex~J Andonian, Yonatan Belinkov, and David Bau. 2023.
\newblock \href {https://openreview.net/forum?id=MkbcAHIYgyS} {Mass-editing memory in a transformer}.
\newblock In \emph{The Eleventh International Conference on Learning Representations}.

\bibitem[{Mozes et~al.(2023)Mozes, He, Kleinberg, and Griffin}]{mozes2023usellmsillicitpurposes}
Maximilian Mozes, Xuanli He, Bennett Kleinberg, and Lewis~D. Griffin. 2023.
\newblock \href {https://arxiv.org/abs/2308.12833} {Use of llms for illicit purposes: Threats, prevention measures, and vulnerabilities}.
\newblock \emph{Preprint}, arXiv:2308.12833.

\bibitem[{OLMo et~al.(2024)OLMo, Walsh, Soldaini, Groeneveld, Lo, Arora, Bhagia, Gu, Huang, Jordan et~al.}]{olmo20242}
Team OLMo, Pete Walsh, Luca Soldaini, Dirk Groeneveld, Kyle Lo, Shane Arora, Akshita Bhagia, Yuling Gu, Shengyi Huang, Matt Jordan, and 1 others. 2024.
\newblock 2 olmo 2 furious.
\newblock \emph{arXiv preprint arXiv:2501.00656}.

\bibitem[{Patil et~al.(2023)Patil, Hase, and Bansal}]{patil2023can}
Vaidehi Patil, Peter Hase, and Mohit Bansal. 2023.
\newblock Can sensitive information be deleted from llms? objectives for defending against extraction attacks.
\newblock \emph{arXiv preprint arXiv:2309.17410}.

\bibitem[{Rafailov et~al.(2023)Rafailov, Sharma, Mitchell, Manning, Ermon, and Finn}]{rafailov2023direct}
Rafael Rafailov, Archit Sharma, Eric Mitchell, Christopher~D Manning, Stefano Ermon, and Chelsea Finn. 2023.
\newblock \href {https://openreview.net/forum?id=HPuSIXJaa9} {Direct preference optimization: Your language model is secretly a reward model}.
\newblock In \emph{Thirty-seventh Conference on Neural Information Processing Systems}.

\bibitem[{Ramakrishna et~al.(2025)Ramakrishna, Wan, Jin, Chang, Bu, Vinzamuri, Cevher, Hong, and Gupta}]{DBLP:journals/corr/abs-2502-15097}
Anil Ramakrishna, Yixin Wan, Xiaomeng Jin, Kai-Wei Chang, Zhiqi Bu, Bhanukiran Vinzamuri, Volkan Cevher, Mingyi Hong, and Rahul Gupta. 2025.
\newblock \href {https://doi.org/10.48550/arXiv.2502.15097} {Lume: Llm unlearning with multitask evaluations}.
\newblock \emph{CoRR}, abs/2502.15097.

\bibitem[{Si et~al.(2023)Si, Zhang, Chang, Zhang, Qu, and Zhang}]{si2023knowledge}
Nianwen Si, Hao Zhang, Heyu Chang, Wenlin Zhang, Dan Qu, and Weiqiang Zhang. 2023.
\newblock Knowledge unlearning for llms: Tasks, methods, and challenges.
\newblock \emph{arXiv preprint arXiv:2311.15766}.

\bibitem[{Sukhbaatar et~al.(2015)Sukhbaatar, Weston, Fergus et~al.}]{sukhbaatar2015end}
Sainbayar Sukhbaatar, Jason Weston, Rob Fergus, and 1 others. 2015.
\newblock End-to-end memory networks.
\newblock \emph{Advances in neural information processing systems}, 28.

\bibitem[{Tian et~al.(2024)Tian, Liang, Cheng, Liu, Wang, Sui, Chen, Chen, and Zhang}]{tian2024forget}
Bozhong Tian, Xiaozhuan Liang, Siyuan Cheng, Qingbin Liu, Mengru Wang, Dianbo Sui, Xi~Chen, Huajun Chen, and Ningyu Zhang. 2024.
\newblock To forget or not? towards practical knowledge unlearning for large language models.
\newblock \emph{arXiv preprint arXiv:2407.01920}.

\bibitem[{Wang et~al.(2024)Wang, Zhang, Xu, Xi, Deng, Yao, Zhang, Yang, Wang, and Chen}]{wang2024detoxifying}
Mengru Wang, Ningyu Zhang, Ziwen Xu, Zekun Xi, Shumin Deng, Yunzhi Yao, Qishen Zhang, Linyi Yang, Jindong Wang, and Huajun Chen. 2024.
\newblock Detoxifying large language models via knowledge editing.
\newblock \emph{arXiv preprint arXiv:2403.14472}.

\bibitem[{Wang et~al.(2025{\natexlab{a}})Wang, Han, Yang, Zhu, Liu, and Sugiyama}]{wang2025towards}
Qizhou Wang, Bo~Han, Puning Yang, Jianing Zhu, Tongliang Liu, and Masashi Sugiyama. 2025{\natexlab{a}}.
\newblock \href {https://openreview.net/forum?id=wUtCieKuQU} {Towards effective evaluations and comparisons for {LLM} unlearning methods}.
\newblock In \emph{The Thirteenth International Conference on Learning Representations}.

\bibitem[{Wang et~al.(2025{\natexlab{b}})Wang, Zhou, Zhou, Shin, Han, and Weinberger}]{wang2025rethinking}
Qizhou Wang, Jin~Peng Zhou, Zhanke Zhou, Saebyeol Shin, Bo~Han, and Kilian~Q Weinberger. 2025{\natexlab{b}}.
\newblock \href {https://openreview.net/forum?id=huo8MqVH6t} {Rethinking {LLM} unlearning objectives: A gradient perspective and go beyond}.
\newblock In \emph{The Thirteenth International Conference on Learning Representations}.

\bibitem[{Wang and Veitch(2024)}]{wang2024does}
Zihao Wang and Victor Veitch. 2024.
\newblock \href {https://openreview.net/forum?id=oZXcwWTCfe} {Does editing provide evidence for localization?}
\newblock In \emph{ICML 2024 Workshop on Mechanistic Interpretability}.

\bibitem[{Wolf et~al.(2020)Wolf, Debut, Sanh, Chaumond, Delangue, Moi, Cistac, Rault, Louf, Funtowicz, Davison, Shleifer, von Platen, Ma, Jernite, Plu, Xu, Le~Scao, Gugger, Drame, Lhoest, and Rush}]{wolf-etal-2020-transformers}
Thomas Wolf, Lysandre Debut, Victor Sanh, Julien Chaumond, Clement Delangue, Anthony Moi, Pierric Cistac, Tim Rault, Remi Louf, Morgan Funtowicz, Joe Davison, Sam Shleifer, Patrick von Platen, Clara Ma, Yacine Jernite, Julien Plu, Canwen Xu, Teven Le~Scao, Sylvain Gugger, and 3 others. 2020.
\newblock \href {https://doi.org/10.18653/v1/2020.emnlp-demos.6} {Transformers: State-of-the-art natural language processing}.
\newblock In \emph{Proceedings of the 2020 Conference on Empirical Methods in Natural Language Processing: System Demonstrations}, pages 38--45, Online. Association for Computational Linguistics.

\bibitem[{Ye et~al.(2022)Ye, Fu, Song, Yang, Liu, Jin, Song, and Wang}]{yeetaal}
Jingwen Ye, Yifang Fu, Jie Song, Xingyi Yang, Songhua Liu, Xin Jin, Mingli Song, and Xinchao Wang. 2022.
\newblock \href {https://doi.org/10.1007/978-3-031-20083-0_6} {Learning with\&nbsp;recoverable forgetting}.
\newblock In \emph{Computer Vision – ECCV 2022: 17th European Conference, Tel Aviv, Israel, October 23–27, 2022, Proceedings, Part XI}, page 87–103, Berlin, Heidelberg. Springer-Verlag.

\bibitem[{Zhang et~al.(2024)Zhang, Lin, Bai, and Mei}]{zhang2024negative}
Ruiqi Zhang, Licong Lin, Yu~Bai, and Song Mei. 2024.
\newblock \href {https://openreview.net/forum?id=MXLBXjQkmb} {Negative preference optimization: From catastrophic collapse to effective unlearning}.
\newblock In \emph{First Conference on Language Modeling}.

\end{thebibliography}

\appendix
\clearpage
\onecolumn
\section{Unlearning Methods Details}
\label{sec:appendixA}
\begin{itemize}
\item \textbf{WGA} \cite{wang2025rethinking} is a reweighted variant of Gradient Ascent (GA) \cite{jang-etal-2023-knowledge} that assigns greater influence to high-confidence tokens by weighting the token-wise log-likelihood using the model’s own predicted probabilities, scaled by a temperature parameter~$\alpha$:
\begin{equation*}
\mathcal{L}_{\text{WGA}} = -\mathbb{E}_{(x, y) \sim \mathcal{D}_{\text{forget}}} \left[ \sum_{i} p_\theta(y_i \mid y_{<i}, x)^\alpha \cdot \log p_\theta(y_i \mid y_{<i}, x) \right],
\end{equation*}
where \(\mathcal{D}_{\text{forget}}\) denotes forget set, and $p_\theta(y_i \mid y_{<i}, x)$ is the predicted probability of the $i$-th token. $\alpha$ is set to 0.1 throughout the experiments.
\item \textbf{DPO} \cite{rafailov2023direct, zhang2024negative} formulates unlearning as a preference optimization problem using paired comparisons between “preferred” and “dispreferred” responses. The model is trained to assign higher likelihood to the preferred output relative to a reference model.
In the context of unlearning, the preferred response corresponds to an “I don’t know” variant, while the dispreferred response is the original answer. The DPO loss is defined as:
\begin{equation*}
    \mathcal{L}_{\text{DPO}} = -\frac{1}{\beta} \cdot \mathbb{E}_{(x, y^{\text{win}}, y^{\text{lose}}) \sim \mathcal{D}_{\text{paired}}} \left[ \log \sigma \left( \beta \cdot \left[ \log \frac{p_\theta(y^{\text{win}} \mid x)}{p_{\text{ref}}(y^{\text{win}} \mid x)} - \log \frac{p_\theta(y^{\text{lose}} \mid x)}{p_{\text{ref}}(y^{\text{lose}} \mid x)} \right] \right) \right],
\end{equation*}
where $\sigma(\cdot)$ is the sigmoid function, $\beta$ is the inverse temperature parameter, and \( p_{\text{ref}} \) denotes the predicted probability computed by the reference model. $\beta$ is set to 0.5 throughout the experiments.
\item \textbf{NPO} \cite{zhang2024negative} extends DPO to the unlearning setting by removing the need for positive (preferred) responses. Each example in the forget set is treated as a negative-only preference signal, encouraging the model to assign lower likelihood to forget data compared to a fixed reference model. Formally, the NPO loss drops the positive term from DPO and becomes:
\begin{equation*}
\mathcal{L}_{\text{NPO}} = -\frac{2}{\beta} \cdot \mathbb{E}_{(x, y) \sim \mathcal{D}_{\text{forget}}} \left[ \log \sigma \left( -\beta \cdot \log \frac{p_\theta(y \mid x)}{p_{\text{ref}}(y \mid x)} \right) \right],
\end{equation*}
where the notation follows that of DPO.  $\beta$ is set to 0.5 throughout the experiments.

\item \textbf{RMU} \cite{li2024the} aims to degrade the internal representations of target knowledge by pushing the hidden states of forget examples toward a fixed random direction. Specifically, a random unit vector $u$ is sampled uniformly from $[0,1)$ and held fixed throughout training. For each forget example, the model is trained to align its hidden states toward $c \cdot u$, where $c$ is a scaling factor. The RMU loss is defined as:
\begin{equation*}
    \mathcal{L}_{\text{RMU}} = \mathbb{E}_{x \sim \mathcal{D}_{\text{forget}}} \left[ \frac{1}{|x|} \sum_{t \in x} \left\| h_\theta^{(\ell)}(t) - c \cdot u \right\|_2^2 \right],
\end{equation*}
where $h_\theta^{(\ell)}(t)$ denotes the hidden state at token $t$ from layer $\ell$, and $|x|$ is the token length of $x$. We use $\ell = 21$ and $c = 2$ in all experiments.

\end{itemize}
\clearpage
\onecolumn
\section{Evaluation Details}
\label{sec:appendixB}
\begin{figure}
    \centering
    \includegraphics[scale=0.3]{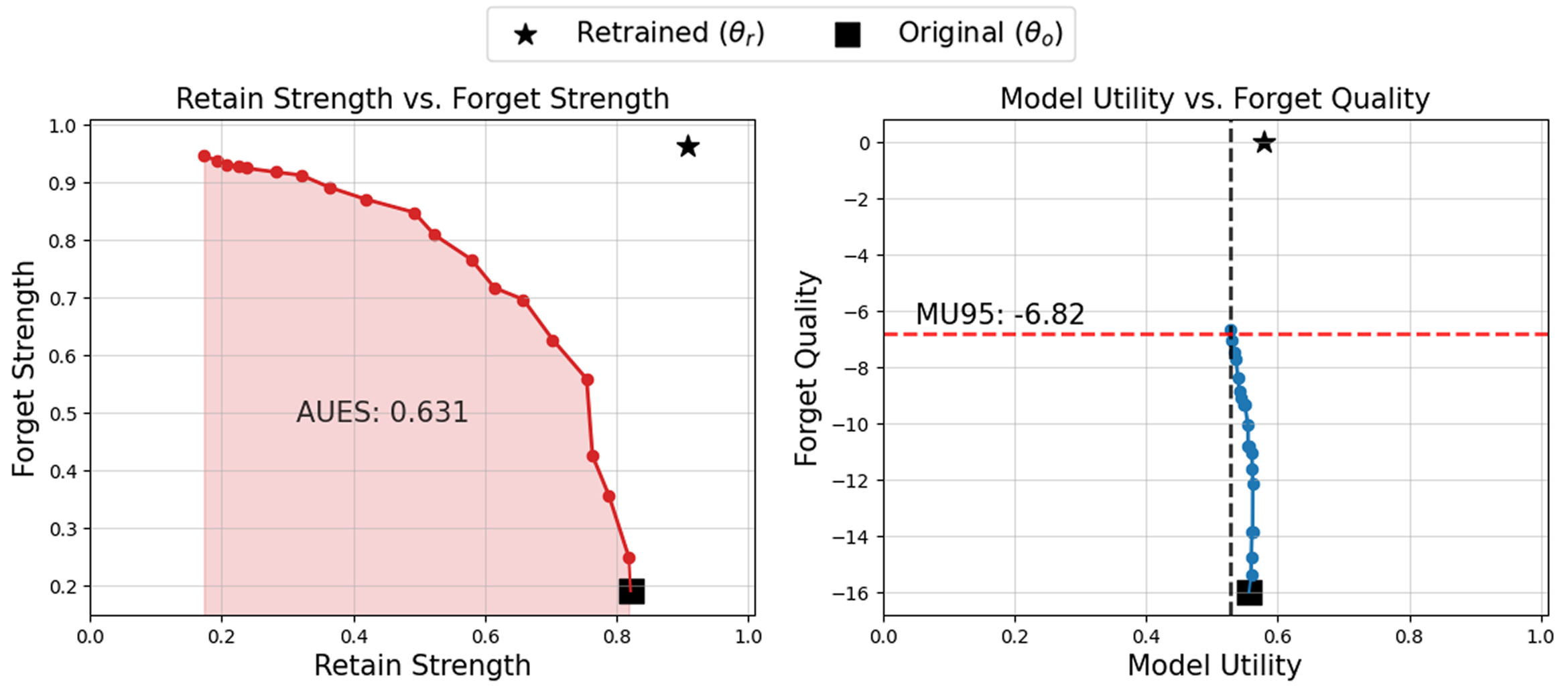}
    \caption{An illustrative example of how AUES and MU95 are calculated from each curve.}
    \label{fig:appendixb}
\end{figure}
\begin{itemize}
    \item \textbf{Extraction Strength Computation} \\
    Following~\citet{wang2025towards}, we quantify the strength of memorization using Extraction Strength (ES), defined as the minimum fraction of the output prefix required to recover the suffix. Formally,
    \[
    \text{ES}(x, y; \theta) = 1 - \frac{1}{|y|} \min_k \left\{ k \;\middle|\; f([x, y_{<k}]; \theta) = y_{\geq k} \right\},
    \]
    where $f$ denotes the model’s output, $x$ is the input, and $y$ is the target output, with $y_{<k}$ and $y_{\geq k}$ representing the prefix and suffix of $y$ split at position $k$. Accordingly, Forget Strength (FS) captures the reduction in ES on the forget set and is computed as \( \text{FS} = 1 - \mathbb{E}_{(x, y) \sim \mathcal{D}_{\text{forget}}} [ \text{ES}(x, y; \theta) ] \). In contrast, Retain Strength (RS) reflects the preserved memorization over the retain set, defined as \( \text{RS} = \mathbb{E}_{(x, y) \sim \mathcal{D}_{\text{retain}}} [ \text{ES}(x, y; \theta) ] \), where \( \mathcal{D}_{\text{retain}} = \mathcal{D} \setminus \mathcal{D}_{\text{forget}} \).

        
\item \textbf{AUES}\\
As discussed, AUES is computed as the area under the FS–RS curve, where each point on the curve corresponds to a pair of FS and RS values achieved under a particular unlearning intensity. To obtain a diverse range of such points across varying extents of unlearning, we adopt a flexible post-unlearning control technique known as \textit{model mixing}~\cite{wang2025towards}. Model mixing allows us to control the extent of unlearning by interpolating between two models: the unlearned model and original (pre-unlearning) model. By mixing the parameters from these two models, the resulting model inherits properties from both—akin to a model ensemble—thereby enabling fine-grained adjustment of unlearning strength.

Specifically, we first unlearn the model to an extent such that FS approaches 1.0 (but does not fully reach it) to avoid collapsing the model entirely, achieved solely by controlling the learning rate. The resulting unlearned model \( \theta \) is then linearly interpolated with the original model \( \theta_o \) using a mixing coefficient \( \alpha \in [0, 1] \), yielding the interpolated parameters:
\[
(1 - \alpha) \cdot \theta_o + \alpha \cdot \theta,
\]
By sweeping \( \alpha \) from 0 to 1 in steps of 0.05, we obtain multiple intermediate models with varying unlearning strengths, allowing us to construct a smooth FS–RS curve and compute AUES reliably. 

\item \textbf{MU95} \\
Similar to AUES, we also leverage model mixing to compute MU95. By generating intermediate models with varying degrees of unlearning through the same interpolation process, we obtain a set of points forming the MU–FQ curve, where each point represents the trade-off between Model Utility (MU) and Forget Quality (FQ) at a specific unlearning strength.

MU95 is then defined as the FQ value observed when MU drops to 95\% of the original model's performance. That is, we interpolate the MU–FQ curve to find the point at which MU reaches 95\% of the initial MU (i.e., the MU of \( \theta_o \)), and take the corresponding FQ at that point.

\end{itemize}

\section{Localization Methods Details}
\label{sec:appendixC}
In this section, we describe the details of the localization methods considered in Section~\ref{revisit}. To formalize, each method assigns an attribution score \( \mathcal{A}^{\ell}(i) \) to each value vector \( \mathbf{v}_i^{\ell} \), the \( i \)-th value vector in the \( \ell \)-th layer, with respect to a given input \( x \), quantifying the extent to which it carries knowledge relevant to processing the input.
\begin{itemize}
\item \textbf{Activation}~\cite{chang-etal-2024-localization} is motivated by the key-value memory theory established by~\citet{geva-etal-2022-transformer}, which suggests that each concept (or piece of knowledge) encoded within the value vectors is promoted and integrated into the \textit{residual stream} through its corresponding memory coefficient (see Section \ref{sec:keyvalue}). Accordingly, Activation simply uses the magnitude of this coefficient as a proxy for the contribution of each value vector.

Formally, the attribution score \( \mathcal{A}^\ell(i) \) is defined as the average over the suffix length \( T \) of the product between the absolute activation coefficient and the norm of the corresponding value vector:
\[
\mathcal{A}^\ell(i) = \frac{1}{T} \sum_{t=1}^{T} \left| h_{i, t}^\ell \right| \left\| \mathbf{v}_{i}^{\ell} \right\|,
\]
where \( h_{i, t}^\ell \) denotes the activation coefficient of the \( i \)-th value vector at layer \( \ell \) at timestep \( t \), when the input consists of all tokens preceding position \( t \), i.e., \( [p, s_{<t}] \). As an additional step, we normalize the attribution scores within each layer to enable localization across layers, rather than within each layer independently.

\item \textbf{MemFlex}~\cite{tian2024forget} assigns attribution scores based on how strongly each value vector responds to a perturbation. Specifically, given a forget example \( (x_u, y_u) \), the label is randomly replaced with \( y_u^* \), and the resulting gradient \( \nabla_\theta \mathcal{L}(x_u, y_u^*; \theta_o) \) is computed. This process is repeated multiple times and averaged to obtain a stable unlearning gradient \( \mathbf{g}^{\ell, \text{unl}}_i \) for \( \mathbf{v}_i^\ell \). The same procedure is applied to the retain set to obtain a retention gradient \( \mathbf{g}^{\ell, \text{ret}}_i \).

Each value vector \( \mathbf{v}_i^\ell \) is then scored based on the direction and magnitude of these gradients. Formally, the attribution score is given by:
\[
\mathcal{A}^\ell(i) = \mathbbm{1} \left[ \cos\left( \mathbf{g}^{\ell, \text{unl}}_i, \mathbf{g}^{\ell, \text{ret}}_i \right) < \mu \;\land\; \left\| \mathbf{g}^{\ell, \text{unl}}_i \right\| > \sigma \right],
\]
where \( \cos(\cdot, \cdot) \) denotes the cosine similarity, \( \mu \) and \( \sigma \) are thresholds for cosine similarity and gradient magnitude, respectively. We controlled \( \mu \) and \( \sigma \) such that approximately 10\% of the value vectors are selected. Specifically, \( \mu \) was set to 0.95, while \( \sigma \) was set to \( 1.6 \times 10^{-4} \) for LLaMA3.1-8B-Instruct and \( 1.4 \times 10^{-4} \) for OLMo2-7B-Instruct. 

\item \textbf{WAGLE}~\cite{jia2024wagle} scores each value vector based on its contribution to forgetting while penalizing its potential interference with retention. While the original WAGLE method scores individual parameters, we aggregate these scores by averaging over the parameters within each  \( \mathbf{v}_i^\ell \), treating the result as its attribution score.

Formally, the attribution score for \( \mathbf{v}_i^\ell \) is computed as:
\[
\mathcal{A}^\ell(i) = \frac{1}{|\mathbf{v}_i^\ell|} \sum_{j \in \mathbf{v}_i^\ell} \left( \left[ \theta_o \right]_j \left[ \nabla \mathcal{L}_{\text{f}}(\theta_o) \right]_j - \frac{1}{\gamma} \left[ \nabla \mathcal{L}_{\text{r}}(\theta_o) \right]_j \left[ \nabla \mathcal{L}_{\text{f}}(\theta_o) \right]_j \right),
\]
where the index set \( j \in \mathbf{v}_i^\ell \) refers to the parameters belonging to \( \mathbf{v}_i^\ell \), and \( \gamma \) is an empirical scaling factor estimated as the average diagonal Hessian value over the retain set.
\end{itemize}
\section{Statistical Significance Testing Details}
\label{sec:appendixD}

To test the statistical significance of the observed differences in AUES and MU95 between two unlearning scenarios, we use non-parametric tests specifically designed for each metric.
\paragraph{AUES permutation test}  
The null hypothesis (\( H_0 \)) is that the two scenarios yield AUES values drawn from the same underlying distribution, that is, there is no meaningful difference in their ability to trade off forgetting and retaining. Under this assumption, the pairing of FS–RS values with their original scenario labels is arbitrary and exchangeable.

To test this, we first compute the observed absolute difference between the AUES values of the two scenarios. Then, for each permutation round, we randomly swap the paired FS–RS points between the two groups with 50\% probability for each value of \( \alpha \), the model mixing coefficient. We recompute the AUES for each permuted group and record the absolute difference. Repeating this process over many iterations yields an empirical null distribution of AUES differences under \( H_0 \).

The p-value is then computed as the proportion of permutations in which the permuted difference equals or exceeds the observed one. A small p-value (e.g., \( p < 0.05 \)) indicates that the observed AUES difference is unlikely to have occurred by chance, thus providing evidence against the null hypothesis.

\paragraph{MU95 bootstrap test}  
The null hypothesis (\( H_0 \)) is that there is no significant difference in MU95 between the two unlearning scenarios, that is, both scenarios exhibit similar forgetting--retention trade-offs at the fixed MU threshold.

To test this, we compute the observed absolute difference in MU95 between the two scenarios. Then, we combine the MU--FQ points from both scenarios into a single pool and perform bootstrap resampling: in each round, we randomly shuffle and split the pooled points into two groups of the original sizes. For each resampled group, we identify the FQ value at the point where MU reaches 95\% and compute the absolute difference in MU95 between the two groups.

This process is repeated over many iterations to construct an empirical null distribution of MU95 differences under \( H_0 \). The p-value is then calculated as the proportion of bootstrap samples where the resampled MU95 difference is greater than or equal to the observed difference. A low p-value (e.g., \( p < 0.05 \)) suggests that the observed MU95 difference is unlikely to have occurred by chance, providing evidence against the null hypothesis.


\section{Supplementary Experiments}
\label{sec:appendixE}

\subsection{Section~\ref{sec:preliminaries} Experiment}
To test whether our result generalizes beyond the original setup (NPO objective with the top 10\% of ranked vectors), we conducted three follow-up experiments, each modifying a single aspect of the main configuration. Specifically, (i) we kept the NPO objective and reduced the update fraction from the top 10\% to the top 5\% of ranked vectors; (ii) we retained the top 10\% selection but replaced the NPO objective with RMU; and (iii) we faithfully reproduced WAGLE's original individual-weight localization procedure. 
All experiments in this appendix use LLaMA3.1-8B-Instruct and are averaged over three random seeds, reported as mean with the standard deviation shown as a subscript.

\begin{table}[t]
\centering
\small
\renewcommand{\arraystretch}{1.2}
\setlength{\tabcolsep}{0.6em}
\begin{tabular}{lcc}
\toprule
\textbf{Method} & \textbf{AUES} & \textbf{MU95} \\
\midrule
Random & $0.521_{\pm \text{0.004}}$ & $\textbf{\text{-}14.34}_{\pm \text{0.48}}$ \\
WAGLE  & $\textbf{0.526}$             & $\text{-}16.82$            \\
\bottomrule
\end{tabular}
\caption{AUES and MU95 at a 5\% localization ratio under the NPO objective.}
\label{tab:supp_5pct_npo}
\end{table}

\paragraph{Varying the localization ratio (5\% under NPO)}
As shown in Table~\ref{tab:supp_5pct_npo}, WAGLE attains a slightly higher AUES than Random, but the improvement is at best marginally significant and it exhibits a worse MU95 score. These results indicate that our conclusion holds at a 5\% localization ratio.

\begin{table}[t]
\centering
\small
\renewcommand{\arraystretch}{1.2}
\setlength{\tabcolsep}{0.6em}
\begin{tabular}{lcc}
\toprule
\textbf{Method} & \textbf{AUES} & \textbf{MU95} \\
\midrule
Random & $0.493_{\pm \text{0.003}}$ & $\textbf{\text{-}18.95}_{\pm \text{0.28}}$ \\
WAGLE  & $\textbf{0.498}$             & $\text{-}19.15$            \\
\bottomrule
\end{tabular}
\caption{AUES and MU95 under the RMU objective with a 10\% localization ratio.}
\label{tab:supp_rmu_10pct}
\end{table}

\paragraph{Changing the unlearning objective (RMU with 10\%)}
Table~\ref{tab:supp_rmu_10pct} shows that under the RMU objective WAGLE again yields only a marginal AUES increase over Random, while MU95 remains worse. This corroborates that our finding is consistent across different unlearning objectives.

\begin{table}[t]
\centering
\small
\renewcommand{\arraystretch}{1.2}
\setlength{\tabcolsep}{0.6em}
\begin{tabular}{lcc}
\toprule
\textbf{Method} & \textbf{AUES} & \textbf{MU95} \\
\midrule
Random & $\textbf{0.536}_{\pm \text{0.001}}$ & $\textbf{\text{-}11.40}_{\pm \text{0.02}}$ \\
WAGLE  & $0.532$             & $\text{-}13.54$            \\
\bottomrule
\end{tabular}
\caption{AUES and MU95 for individual-weight localization.}
\label{tab:supp_wagle_weights}
\end{table}

\paragraph{Reproducing WAGLE’s original procedure (individual-weight localization)}
We precisely reproduced WAGLE’s original setup, which assigns an attribution score to each individual weight rather than to MLP value vectors. Following \citet{jia2024wagle}, we performed localized unlearning under the NPO objective, applying updates to the top 80\% of weights by attribution score and comparing it with localized unlearning on 80\% of randomly selected weights. As summarized in Table~\ref{tab:supp_wagle_weights}, WAGLE yields a slightly lower AUES than Random and a worse MU95, further confirming that our conclusion holds even when localization targets individual weights rather than value vectors.

\paragraph{Summary}
Across all three follow-up experiments: (i) reducing the localization ratio to 5\% under NPO, (ii) switching the unlearning objective to RMU at 10\%, and (iii) reproducing WAGLE’s original individual-weight localization, the purported advantage of WAGLE over Random is at best marginal in AUES and is consistently accompanied by worse MU95. These results reinforce that our main finding holds across localization ratios, unlearning objectives, and localization target units.

\subsection{Section~\ref{controlled_exp} Experiment}
To examine whether our findings generalize beyond TOFU, we evaluated the Oracle vs.\ Random comparison on the LUME Task~2 PII dataset \cite{DBLP:journals/corr/abs-2502-15097}, which contains 2{,}025 AI-generated synthetic personal records (e.g., phone numbers, e\mbox{-}mail addresses) designed for forgetting fictitious PIIs. The synthetic nature of LUME, as with TOFU, allows us to precisely control which parameter regions encode the target knowledge and removes the possibility of prior exposure during pretraining.

\paragraph{Setup}
We used a 10\% forget ratio, a 10\% localization ratio, and the NPO objective, averaging over three random seeds. Because Forget Quality is only defined on TOFU by design, we report AUES only for LUME.

\begin{table}[t]
\centering
\small
\renewcommand{\arraystretch}{1.2}
\setlength{\tabcolsep}{0.8em}
\begin{tabular}{lcccc}
\toprule
\textbf{Metric} & \textbf{Random} & \textbf{Oracle} & $|\Delta|$ & $p$-val \\
\midrule
AUES & $0.473_{\pm \text{0.028}}$ & $\textbf{0.482}_{\pm \text{0.017}}$ & $0.017$ & 0.44 \\
\bottomrule
\end{tabular}
\caption{AUES on the LUME Task~2 PII dataset with a 10\% forget ratio and a 10\% localization ratio under the NPO objective. Values are reported as mean with the standard deviation shown as a subscript, computed over three random seeds.}
\label{tab:supp_lume}
\end{table}

\paragraph{Results}
As shown in Table~\ref{tab:supp_lume}, Oracle and Random achieve nearly identical AUES, differing by only $0.017$ ($p{=}0.44$), which is statistically negligible. This mirrors our TOFU results and further indicates that successful localization does not necessarily translate into improved unlearning performance on LUME.

\section{Hyperparameter Details}
\label{sec:appendixF}

\paragraph{Global settings}
We use a batch size of $16$, a weight decay of $0.01$, and train for $5$ epochs in all experiments. Method-specific hyperparameters for each unlearning objective and localization method are detailed in Appendix~\ref{sec:appendixA} and Appendix~\ref{sec:appendixC}, respectively.

\paragraph{Model checkpoints}
We initialize from the publicly released instruction-tuned checkpoints on Hugging Face:
\href{https://huggingface.co/meta-llama/Llama-3.1-8B-Instruct}{LLaMA3.1--8B--Instruct}
(\texttt{meta-llama/Llama-3.1-8B-Instruct}) and
\href{https://huggingface.co/allenai/OLMo-2-1124-7B-Instruct}{OLMo2--7B--Instruct}
(\texttt{allenai/OLMo-2-1124-7B-Instruct}).


\begin{tcolorbox}[acad/base, title={Section~\ref{sec:preliminaries} Experiments}]
\begin{itemize}[leftmargin=*, itemsep=5pt, topsep=3pt, parsep=0pt]
  \item \textbf{Learning Full Data:} LR $9.5\mathrm{e}^{-6}$.

  \item \textbf{Unlearning:}
  \begin{itemize}[leftmargin=1.6em, itemsep=4pt, topsep=2pt, parsep=0pt]
    \item \textbf{Mask Seeds:} $\{7, 19, 99\}$.
    \item \textbf{Learning Rates:}
    \begin{itemize}[leftmargin=1.4em, itemsep=3pt, topsep=1.5pt, parsep=0pt]
      \item \textbf{Table ~\ref{tab:preliminary}:} Random $6\mathrm{e}^{-5}$; Activations $1\mathrm{e}^{-5}$; WAGLE $1\mathrm{e}^{-5}$; MemFlex $4\mathrm{e}^{-5}$.
      \item \textbf{Table ~\ref{tab:supp_5pct_npo}:} Random and WAGLE $1\mathrm{e}^{-4}$.
      \item \textbf{Table ~\ref{tab:supp_rmu_10pct}:} Random $2\mathrm{e}^{-5}$; WAGLE $8\mathrm{e}^{-6}$.
      \item \textbf{Table ~\ref{tab:supp_wagle_weights}:} Random and WAGLE $1\mathrm{e}^{-5}$.
    \end{itemize}
  \end{itemize}
\end{itemize}
\end{tcolorbox}


\begin{tcolorbox}[acad/base, title={Section~\ref{controlled_exp} Experiments}]
\begin{itemize}[leftmargin=*, itemsep=5pt, topsep=3pt, parsep=0pt]

  \item \textbf{Table ~\ref{tab:main}}
  \begin{itemize}[leftmargin=1.6em, itemsep=4pt, topsep=2pt, parsep=0pt]
    \item \textbf{Mask Seeds:} $\{7, 19, 31, 47, 99\}$.
    \item \textbf{LLaMA3.1-8B-Instruct}
    \begin{itemize}[leftmargin=1.4em, itemsep=3pt, topsep=1.5pt, parsep=0pt]
      \item \textbf{Learning:} Retain set LR $1\mathrm{e}^{-5}$; Forget set LR $2\mathrm{e}^{-4}$ with $\lambda_{\mathrm{retain}}=2$.
      \item \textbf{Unlearning (Learning Rates):} WGA $2\mathrm{e}^{-5}$; NPO $8\mathrm{e}^{-5}$; DPO $4\mathrm{e}^{-4}$; RMU $2.3\mathrm{e}^{-5}$.
    \end{itemize}
    \item \textbf{OLMo2-7B-Instruct}
    \begin{itemize}[leftmargin=1.4em, itemsep=3pt, topsep=1.5pt, parsep=0pt]
      \item \textbf{Learning:} Retain set LR $1\mathrm{e}^{-5}$; Forget set LR $3\mathrm{e}^{-4}$ with $\lambda_{\mathrm{retain}}=2$.
      \item \textbf{Unlearning (Learning Rates):} WGA $2\mathrm{e}^{-5}$; NPO $5\mathrm{e}^{-5}$; DPO $1.5\mathrm{e}^{-3}$; RMU $3\mathrm{e}^{-5}$.
    \end{itemize}
  \end{itemize}
  
  \item \textbf{Figure ~\ref{fig:discussion}}
  \begin{itemize}[leftmargin=1.6em, itemsep=3pt, topsep=2pt, parsep=0pt]
    \item \textbf{Mask Seeds:} Oracle $=7$; Random A $=7$, B $=11$, C $=49$.
    \item \textbf{Learning:} Same settings as above.
    \item \textbf{Unlearning:} LR $4\mathrm{e}^{-4}$ with $\alpha=2$.
  \end{itemize}
  
  \item \textbf{Table ~\ref{tab:supp_lume}}
  \begin{itemize}[leftmargin=1.6em, itemsep=3pt, topsep=2pt, parsep=0pt]
    \item \textbf{Learning:} Retain set LR $3\mathrm{e}^{-5}$; Forget set LR $3\mathrm{e}^{-4}$ with $\lambda_{\mathrm{retain}}=2$.
    \item \textbf{Unlearning:} LR $2\mathrm{e}^{-4}$.
  \end{itemize}

\end{itemize}
\end{tcolorbox}

\section{Resources, Licenses, and Packages}
\label{sec:appendixG}
We used the TOFU dataset \cite{maini2024tofu}, which is released under the MIT License. All experiments were conducted on two NVIDIA A100 GPUs with 80\,GB of VRAM each. We relied on several publicly available libraries, including \texttt{transformers} \cite{wolf-etal-2020-transformers} and \texttt{datasets} \cite{lhoest-etal-2021-datasets}. We made use of AI assistants, specifically ChatGPT, to help with code implementation and to aid in writing.
\end{document}